\documentclass[pdflatex,sn-mathphys-num]{sn-jnl}

\usepackage{graphicx}%
\usepackage{multirow}%
\usepackage{amsmath,amssymb,amsfonts}%
\usepackage{amsthm}%
\usepackage{mathrsfs}%
\usepackage[title]{appendix}%
\usepackage{xcolor}%
\usepackage{bm}%
\usepackage{textcomp}%
\usepackage{manyfoot}%
\usepackage{booktabs}%
\usepackage{algorithmic}
\usepackage{algorithm}%
\usepackage{subfigure}
\usepackage{multirow}
\usepackage{listings}%
\newcommand{\tabincell}[2]{\begin{tabular}{@{}#1@{}}#2\end{tabular}}

\theoremstyle{thmstyleone}%
\theoremstyle{thmstyletwo}%

\theoremstyle{thmstylethree}%

\begin{document}

\title[ Soft causal learning for generalized molecule property prediction: An environment perspective]{Soft causal learning for generalized molecule property prediction: An environment modeling perspective}


\author[1]{Limin Li}\email{lilimin@mail.ustc.edu.cn}
\author[2]{Kuo Yang}\email{yangkuo@mail.ustc.edu.cn}
\author[1,2]{Wenjie Du}\email{duwenjie@ustc.edu.cn}
\author[1,2]{Pengkun Wang}\email{pengkun@ustc.edu.cn}
\author*[1,2]{Zhengyang Zhou}\email{zzy0929@ustc.edu.cn}
\author[1,2]{Yang Wang}\email{angyan@ustc.edu.cn}

\affil[1]{\orgdiv{School of Software Engineering}, \orgname{University of Science and Technology of China}, \orgaddress{\street{Jinzhai Road}, \city{Hefei}, \postcode{230000}, \state{Anhui}, \country{China}}}

\affil[2]{\orgdiv{Suzhou Institute for Advanced Research}, \orgname{University of Science and Technology of China}, \orgaddress{\street{Renai Road}, \city{Suzhou}, \postcode{215000}, \state{Jiangsu}, \country{China}}}


\abstract{Learning on molecule graphs has become an increasingly important topic in AI for science, which takes full advantage of AI to facilitate scientific discovery. Existing solutions on modeling molecules utilize  Graph Neural Networks (GNNs) to achieve  representations but they mostly fail to adapt models to out-of-distribution (OOD) samples. Although recent advances on OOD-oriented graph learning have discovered the invariant rationale on graphs, they still ignore three important issues, i.e., 1) the expanding atom patterns regarding  environments on graphs lead to failures of invariant rationale based models, 2) the associations between discovered molecular subgraphs and corresponding properties  are complex where causal substructures cannot fully interpret the labels. 3) the interactions between environments and invariances can influence with each other thus are challenging to be modeled. To this end, we propose a soft causal learning framework, to tackle the unresolved OOD challenge in molecular science, from the perspective of fully modeling the molecule environments and bypassing the invariant subgraphs. Specifically, we first incorporate chemistry theories into our graph growth generator to imitate expaned environments, and then devise an GIB-based objective to disentangle environment from whole graphs and finally introduce a cross-attention based soft causal interaction, which allows dynamic interactions between environments and invariances. We perform experiments on seven datasets by imitating different kinds of OOD generalization scenarios. Extensive comparison, ablation experiments as well as visualized case studies demonstrate well generalization ability of our proposal.}




\maketitle

Learning on molecules has increasingly become a powerful research topic to enable various applications from molecular property estimation, drug discovery to molecule retrosynthesis~\cite{barbatti2014newton,du2023fusing}, hence benefiting the community of scientific computing. However, molecule properties are mostly tested by labor-intensive experiments with the risk of poisonousness, while the drug discovery process usually costs numerous trial and errors. To this end, how to ensure the efficiency of implementing both academia and industry experiments and maximumly exploiting the power of data intelligence for practical biology and chemistry production become the central attention of researchers.

There have been numerous efforts of various Graph Neural Networks (GNNs). Technically, graph learning on molecular science either focus on finding support invariant substructures for property estimation~\cite{wang2023brave,yang2023extract,wu2022discovering}, or exploring the homophily and heterophily on graphs to improve the representation capacity for final classification~\cite{zheng2022graph}. However, given the explosions of emerging materials~\cite{yi2024achieving} and diversity of biological medicine~\cite{bernstein2014biological}, molecular science suffers the inherent insufficiency of the training sets for model learning. Therefore, learning OOD generalizations on molecular graphs becomes the core obstacle for GNN-based molecular research towards material-based industrial practices and further advances. 

Recent efforts have been made to construct a series of OOD-oriented learning frameworks~\cite{wang2023brave,wu2022discovering,zhao2022learning}. These solutions can be divided into three aspects, i.e., finding invariant substructure rationales~\cite{wu2022discovering}, counterfactual-based data augmentation~\cite{zhao2022learning} as well as the environment augmentations~\cite{chen2023does,yuan2024environment,xia2024deciphering}. Specifically, pioneering research devises a dual optimization strategy, which allows the joint condensation on content subgraph and neural structures~\cite{wang2022searching}. 
\begin{figure}[ht]	
	\centering
	\includegraphics[width=\linewidth]{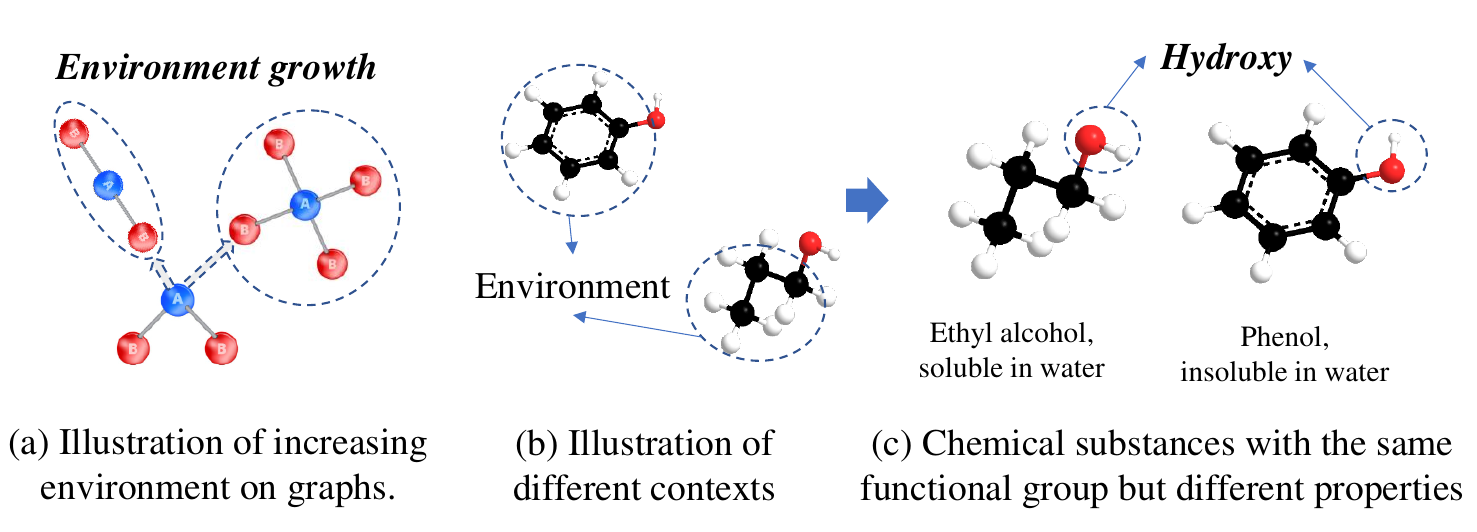}
	\caption{Motivation of CauEMO. (a) Increasingly growing environments can finally dominate the property of whole graph. (b) Two substances, ethyl alcohol and phenol are with the same functional group of hydroxyl, but are with different environmental substructures connection, resulting in different solubility properties.}
	\label{fig:motiva}
\end{figure} 
Indeed,  a graph is widely acknowledged that it can be decomposed into environments and causal invariant rationales. However, with the scale and diversity of molecular graphs increasing, there are two significant issues hindering the generalization of existing OOD solutions~\cite{wu2022discovering,wang2022searching,zhao2022learning,wang2023brave,sun2022graph}, as shown in Fig.~\ref{fig:motiva}. 
First, the increasing types of molecules lead to expanding patterns of environments on graphs, as illustrated in Fig.~\ref{fig:motiva}(a) where the environmental information can dominate the entire graph, resulting in the failure of invariant rationale based models. Second, the associations between discovered molecular subgraphs and corresponding properties (i.e., labels) are complex. Given a molecule graph with specific functional groups, the deterministic property (e.g., solubility) is not fully dependent on the specific functional groups but may partially rely on the environmental substructures around the functional group, which has been demonstrated in Fig.~\ref{fig:motiva}(b). From the perspective of information theory, the information encapsulated in invariant subgraphs is incomplete to interpret labels, and the reason lies in that environments usually have interactive effects with invariant rationales. Therefore, only modeling invariant rationales could lead the model trapping into suboptimal results. A potential solution towards more powerful generalization capacity is to maximize the informativeness by exploiting the auxiliary environments and coordinating environment-invariance. Unfortunately, most existing works focus on subgraph extraction from invariance perspective~\cite{wu2022discovering,yang2023extract}, where very few literature decouples environmental parts and dissects how to couple extracted invariance with environmental substructures. Given the expanding molecules and unlimited synthetic materials, boosting the representation of graph environments and model-aware informativeness for generalization improvement is still faced with two specific challenges.   
\begin{itemize}
	\item The environment patterns are diverse and variable, making it challenging to identify and separate environments.
	\item With environments and label-interpreted part well obtained, how to cooperatively exploit the environment and invariant signals to achieve predictions?
\end{itemize}

To address the above challenges, we propose a Molecular Property prediction network named CauEMO, to systematically tackle the OOD issue. Firstly, to promote the diversity of environments on graphs and thus alleviating the potential dominance of environments in predictions, we design a knowledge-enhanced environment growth generator to simulate the environments for diversity expansion. Secondly, to improve quality of environmental representation, we treat the environment as a mediating variable, and explicitly extract such representation via constructing an Environment-Graph Information Bottleneck learning objective to disentangle label-irrelevant environmental signals and label-relevant signals, allowing sufficient environment squashing. Lastly, given the potential interactions between rationales and environments and limited interpretability between rationales and labels, it is difficult to obtain a complete view of labels solely relying on invariant information. We then design an Environment-Invariance Soft Causal Interaction, which refines environment and allows information interactions between environments and causal invariance. An environment-invariance cross-attention is introduced to realize adaptive information fusion with soft scores dependent on input features. \textbf{The contributions of our work can be three-fold.}  
\begin{itemize}
	\item We discover two main factors constraining molecular graph generalization capacity. The first is the ever-growing and expanding environmental signals on graphs gradually suppressing primary information, and the second is the potential interactions between causal invariance and environments. We then  propose an environment-oriented solution to increase graph diversity and capture environment-rationale interactions to enhance graph representation.
	\item  Technically, bypassing the exploration of invariant subgraphs, we  start the research from environmental modeling. We first explicitly incorporate chemistry principle into our graph growth generator to imitate the environment growth, and introduce a cross-attention soft causal interaction, which allows flexible and dynamic interaction between environments and invariances.
	\item Empirically, we conduct experiments on seven datasets, including two categories of DrugOOD datasets and one synthetic   Motif dataset. These experiments demonstrate the effectiveness of CauEMO, and the practical capacity on generalizing models to unseen graphs with increased environments and designed neural architecture. 
\end{itemize}

\section{Related Work}
\textbf{Graph neural network and subgraph learning.}
Graph neural networks (GNNs) are initially introduced by~\cite{gori2005new} for graph-structured data mining by iteratively aggregating information from neighbors. Recently, with the increasing prosperity of deep learning, GNNs have developed by stacking layers and simplifying the node-level adjacencies to gain powerful representation capacity, such as GCN~\cite{kipf2016semi}, GraphSAGE~\cite{hamilton2017inductive}, and Graph Attention Networks (GAT)~\cite{velivckovic2017graph}, where GAT allows flexible node-level attention. However, conventional deep GNNs usually lack interpretability to explain which specific substructure contributes most to final predictions. To this end, subgraph learning is leveraged to boost the interpretability and generalization. Specifically,  SubGNN~\cite{alsentzer2020subgraph} decouples the graph topology into three property-aware channels to extract subgraph patterns on position, neighborhood, and structure. Furthermore, Sugar 
~\cite{sun2021sugar}, and XGNN~\cite{yuan2020xgnn} devises the reinforcement learning to help extract interpretable subgraphs, and P2GNN~\cite{yang2023extract} is proposed to extract the asymmetric  patterns of substructures in large-scale graphs with considering pivot nodes. However, generalizing models to other unseen scenarios requires capturing invariance across scenarios. Even so, these solutions to graph and subgraph learning fail to explicitly involve invariant factors thus trapped into suboptimal results in most generalization tasks.  

\textbf{Invariant learning for OOD generalization.}
Generalization issues are common in learning-based solutions, ranging from computer vision~\cite{lu2023label,lu2019adaptive}, static graph learning~\cite{wu2022discovering, arjovsky2019invariant,li2022ood} to dynamic graph learning~\cite{du2021adarnn,zhou2023maintaining}. Among them, existing solutions to graph out-of-distribution generalization usually divide the whole graph into environments and invariant rationales~\cite{wu2022discovering}, which is inherited from causal theory~\cite{pearl2009causal}. Representative invariant learning  frameworks~\cite{arjovsky2019invariant,liu2023flood} have been proposed to handle  distribution shifts where it minimizes the summarized risks across different conditions and environments. Following it~\cite{arjovsky2019invariant}, graph-level learning for OOD generalization such as DIR~\cite{wu2022discovering}, OOD-GNN~\cite{li2022ood}, and MoleOOD~\cite{yang2022learning} have been  proposed for molecular scientific research. And an OOD solution on dynamic graph learning CauSTG is devised to capture invariance across sample groups~\cite{zhou2023maintaining}. Recently, EERM~\cite{wu2022handling} overcomes the non-i.d.d. issue on node-level learning and takes a reinforcement learning to enhance the environment diversity. Even though, all these solutions only focus on the invariant factors,  directly ignoring the valuable information within environments. Actually, graph environments can be deemed as the conditions to invariances, where environments are also equipped with valuable information and can potentially interact with invariances to influence label interpretation. Therefore, how to exploit  environments to enhance cooperative learning on both environment and invariant factors for better generalization still remains under-explored.

\textbf{Environment-aware learning for graph OOD generalization.}
Modeling the environments, such as environment representation, generation~\cite{wang2023surban,yuan2024environment}, and augmentation~\cite{xia2024deciphering,zhao2022learning} can be another way to promote OOD capacity. For instance, Zhao, et, al. consider the graph topology as  the virtual environment and devise a counterfactual strategy by imposing perturbation on environments, which can be  viewed as a data augmentation~\cite{zhao2022learning}. CaST employs the back-door adjustment by a novel disentanglement block to separate the temporal environments via structural causal model~\cite{xia2024deciphering}. Moreover, ~\cite{chen2023does} imposes an environment augmentation by introducing an assistant model by maximizing the variations to handle the OOD issue. Besides, ~\cite{yuan2024environment} designs a sampling-generative process to generate new environments while ~\cite{wang2023surban} disentangles the environment representation and imposes an environment-aware contrastive representation learning. Even though, above solutions either construct the closed environment set, or devise IRM and contrastive learning-based strategies to disentangle environments. In fact, the types of molecules are usually becoming more and more diverse  and the number of atoms are increasing. Then the molecules as well as corresponding graph environments  cannot be fully enumerated where the limited extracted invariance cannot fully reflect the summarized properties.  Hence, these above-mentioned solutions fail to mimic the increasing growths of spurious substructures and cannot capture the environment-invariance interactions. 

\textbf{Summary.} Considering  abovementioned graph learning frameworks for molecular science, there are still two significant issues remain unresolved in OOD tasks: 1) the increasing types  and numbers of molecules lead to expanding patterns of environments and  results in the failure of invariant rationale based models. 2) the associations between molecular graphs and corresponding properties are complex while interactions between environment-invariance are intractable to capture. To this end, detouring the invariance and directly enhancing environment modeling can be a promising avenue towards OOD learning improvement. 

\begin{figure*}[ht]	
	\centering
	\includegraphics[width=5.6in]{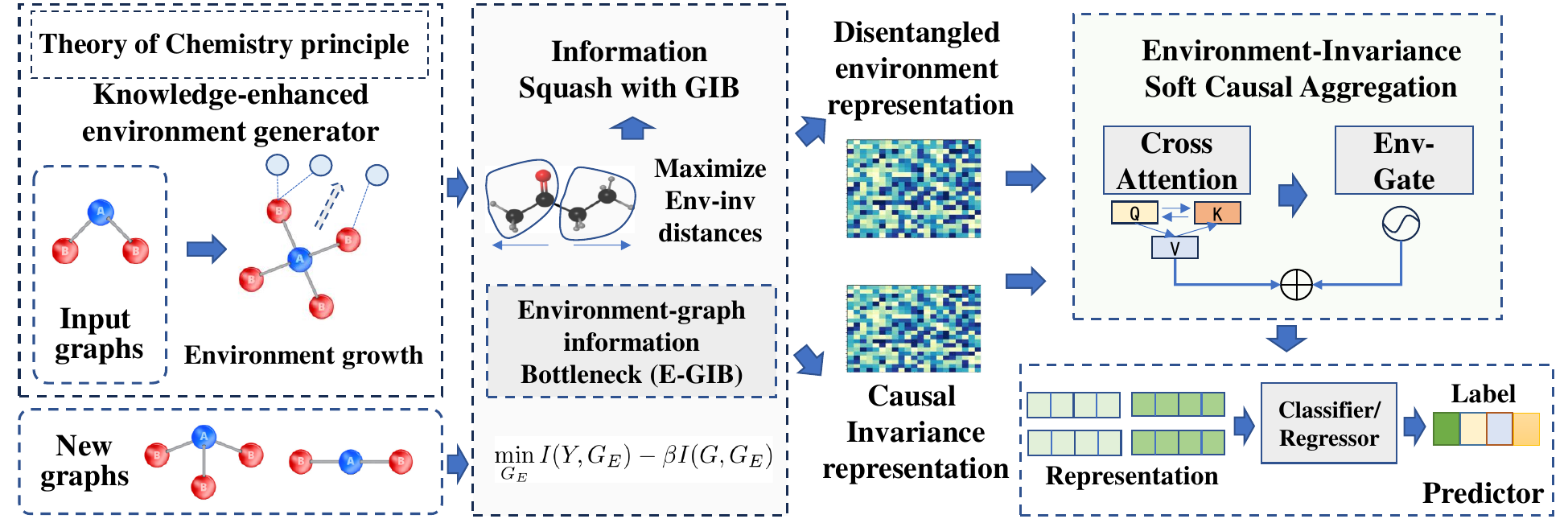}
	\caption{Framework overview of CauEMO}
	\label{fig:FO}
\end{figure*} 
\section{Preliminary and problem definition}
Consider a molecule graph $G=(\mathcal{V}, \mathcal{E})$ where the node  and  edge in $G$  can be denoted as $v_i \in \mathcal{V}$ and $e_{ij}\in \mathcal{E}$. The deterministic observation in node $v_i$ is written as  $x_i\in \bm{X}$, where it shows the representation of atom. Given a graph $G_j$, the molecular science learning is to predict a series of property $\bm{y}_j$, consists of both graph-level regression for continuous property and graph-level classification for categorical properties. 

\textbf{OOD settings.} Given a series of molecular graphs in training set $(\bm{Y}_{tr}, \mathcal{G}_{tr})$ and testing set $(\bm{Y}_{test},  \mathcal{G}_{test})$ where $P(\mathcal{G}_{tr})\ne P(\mathcal{G}_{test})$, we are going to derive a neural function $y=f^*(G)$ with OOD learning capacity that can transfer invariance and adapt new environments to new scenarios.

\section{Methodology}
\subsection{Framework overview}
As shown in Fig.~\ref{fig:FO}, the proposed CauEMO is composed of three well-designed components, i.e., a Knowledge-enhanced environment generator, an E-GIB for irrelevant environment disentanglement and an Environment-Invariance Soft Causal Aggregation (SCA), to respectively imitate the increasing growth  of environments on molecule graphs with chemical domain knowledge, disentangle environment information from whole graphs with information theory guarantee, and dynamically implement environment refinement and environment-invariance interactions for OOD-oriented representation improvement.

\subsection{Knowledge-enhanced environment generator}
The environment, separated from core property substructures, within a molecular graph is also vital for property forecasting, especially for Out-of-Distribution scenarios. The diversity of graph environments can determine the accuracy and epistemic uncertainty of substructure extraction. Unfortunately, existing OOD learning either exploit the disentangled environments within the dataset itself~\cite{wang2024kill}, or explore an augmentation without any constraints of domain knowledge~\cite{yuan2024environment}. In contrast, in our paper,  we devise an environment  generator with the help of domain knowledge, to imitate the growth of environment over molecular graphs and simultaneously maintain the primary principle of chemistry. We first introduce the chemical knowledge-based grouping strategy to decompose the chemical unit into functional group set $\mathcal{G}_I = \{ G_i \}$ and non-deterministic group set $\mathcal{G}_{N}=\{G_n\}$, where the former ones can be seen as the causal invariance for labels while the latter ones are environment sets. In this way, we can dynamically combine the causal invariance with the environment part to increase the environment of graphs by generating substantial new graphs. 
To realize it, we rank the substructures in $G_{n}$ by the number of atoms for imitation of increasing scales of environments, i.e., ${\rm EgoGraph}=\{{s}_1,s_2,\ldots,s_m\}$, we can iteratively replace the $s_k$ in the combined graph. 
However, regarding graph editing task, a serious issue is how to guarantee the rationality of new graphs, i.e., how to ensure the new graphs to satisfy the chemistry principle, this it can be synthesized or bought for industrial chemical engineering. Hence, we introduce the chemical crosslinks to help  judge whether the  substructure connection between two set items are reasonable, where the domain-specific knowledge is interpreted as the law of conservation of charge. Given the connected atom $v_i$ from $\mathcal{G}_I$, we check the summation of its chemical crosslinks. Let us denote the chemical bond of $v_i$ as $d_i$, $v_j$ are a series of neighboring nodes which is going to connected to $v_i$  in set $\mathcal{G}_N$, then we can derive the equation according to the  principle of chemical bond, i.e.,   
\begin{equation}
	{d_i} = \sum\limits_j \in N({v_i}){{p_{ij}}}\Leftrightarrow{v_i}-{v_j}
	\label{eq:equi}
\end{equation}
If such combination $v_i \in \mathcal{G}_I, v_j \in \mathcal{G}_N $ can satisfy above Equation.~(\ref{eq:equi}), we can make the concatenation on the node-level to achieve the new graph, i.e., 
\begin{equation}
	G_0 = G_i, \;\;    G_{new} = {\rm Concat}[G_0; G_j]
\end{equation}

Instead of constructing an environment set from closed training samples without introducing any new molecules~\cite{yuan2024environment,xia2024deciphering}, we especially inherit the chemical domain knowledge, and concatenate the causal invariant part with auxiliary environment part iteratively. To this end, our environment growth generator can increase the uncertainty and diversity of molecules with limited training data and simultaneously ensure fundamental chemistry-specific principle for generating samples, which can work cooperatively with following learning modules for benefiting downstream OOD tasks. 
\begin{figure*}[ht]	
	\centering
	\includegraphics[width=5.4in]{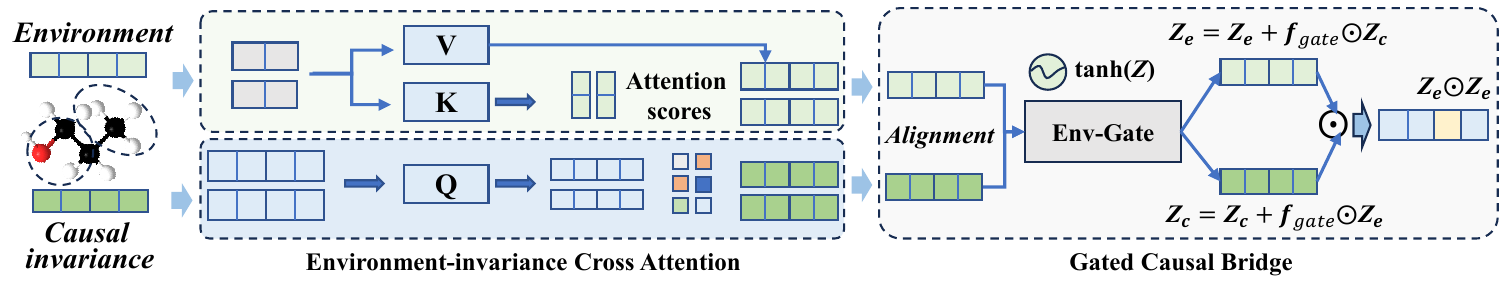}
	\caption{Environment-invariance soft causal interaction}
	\label{fig:OV-SCA}
\end{figure*}

\subsection{E-GIB based irrelevant environment disentanglement}
Most solutions to graph OOD challenges emphasize the causal invariance for transfer. But unfortunately, in real-world molecule-oriented tasks, the invariance across all graphs are limited, in other words, there is limited common invariant parts across all scenarios that can sufficiently support the transfer. In our work, we focus more on the environments, which can be further refined to enhance the improvement and informativeness of causal parts. We thus bypass modeling invariant associations, and propose an Environment-Graph Information Bottleneck (E-GIB) to explicitly extract the environments that are mostly irrelevant with deterministic graph labels, from the perspective of information theory. 

Given a graph $G$, the initialized environment representation $G_E$, and the label $Y$ of $G$, our E-GIB is expected to find out the most label-irrelevant  representation  on graph $G_E$. Our E-GIB will not only squash the environment representation away from labels but also ensures 
the environment $G_E$ can cover most of the graph $G$. By borrowing the theoretical guarantee from information theory, the guided training objective  can be preliminarily described as, 
\begin{equation}
	\mathop {\min }\limits_{{G_E}} I(Y,{G_E}) - \beta I({G},{G_E})
	\label{eq:EGIB}
\end{equation}
where $\beta$ is the hyperparameter setting as 1 according to common practices~\cite{wu2020graph,yu2022improving,yu2020graph}. We will then elaborate the implementation of above E-GIB.

For  the first term in Equation.~(\ref{eq:EGIB}),  we take the standard cross-entropy loss  to instantiate the preliminary objective, which aims to suppress the label-relevant information on graphs by inheriting the tractable lower bound obtained from literature~\cite{yu2020graph}. This learning objective is specified as the environment predictor $P_{\theta}(Y|G_E)$ (a.k.a. $f_\theta$). Regarding the second term, since there is no premise or apriori information  for marginal distribution $p(G_E)$, we derive a variational estimator $\mathbb{Q}(G_E)$ to approximate $p(G_E)$, i.e., $p(G_E) \sim \mathbb{Q}(G_E)$, and obtain the variational upper bound with KL-divergence~\cite{alemi2016deep}. It can be formally derived by,
\begin{equation}
	I({G},G_E) \le {\rm KL} \left( {{P_\phi }( {G|{G_E}})||\mathbb{Q}\left( {{G_E}} \right)} \right)
\end{equation}
Then we can take such KL-divergency to estimate the marginal probability  $p(G_E)$ with parameterized $\mathbb{Q}(G_E;\bm{W}_{ge})$, where $\bm{W}_{ge}$ are learnable parameters to this  probability distribution estimation. We can designate the $P_{\phi}$ (a.k.a. $g_\phi$) as the environment extractor.

\textbf{Learning objective.} Considering both cross-entropy for label-irrelevant information suppression and the KL-divergence for variational estimator, we can obtain the final learning objective for our environment disentanglement, i.e.,
\begin{equation}
	{\rm min} \;\; \mathbb{E}[log \mathbb{P}_\theta(Y|G_E)]-\beta \mathbb{E}[{\rm KL}({P_\phi }({G|{G_E}})||\mathbb{Q}( {G_E})]
\end{equation}

With above learning objective, we can sufficiently disentangle the environment $G_E$ over graph  $G$. Then we can provide the details of how to implement our enviroment extractor and environment predictor.

\textbf{Implementation of environment extractor $g_\phi$.} 
The {environment extractor $g_\phi$} encodes input graph $G$ via GNN and for each edge $(u,v)\in \mathcal{E}$, $g_\phi$ consists of an MLP layer and a sigmoid function that maps the concatenation of two  node representation  into $p_{uv} \in [0,1]$, i.e.,
\begin{equation}
	p_{uv}  =	{\rm MLP}(h_u, h_v; \bm{W}_{g})
\end{equation}
where  $(h_u, h_v)$ are representations of node $u$ and $v$,  $\bm{W}_{g}$ are learnable parameters for environment extractor. For each forward pass during training process, we sample stochastic attention from Bernoul distributions $\alpha_{u,v} \sim {\rm Bern}(p_{uv})$. We also apply the gumbel-softmax reparameterization trick to ensure the continuous gradient for computable $p_{uv}$~\cite{jang2016categorical}.  Then the extracted graph $G$ will have an attention-based selected subgraph as,
\begin{equation}
	\bm{A}_S = \alpha \cdot \bm{A}
\end{equation}
Therefore, the environment subgraph extractor falls into a Stochastic Attention mechanism controlled by Bern distribution.

\textbf{Implementation of environment predictor $f_\theta$.}
The predictor $f_\theta$ adopts the same GNN to encode the extracted graph $G$ to a graph representation and finally passes such representation through an MLP laye plus softmax to model the distribution of $Y$. This procedure enables  the variational distribution $P_\theta(Y|G_E)$.

Finally, the Marginal Distribution can be  Controlled via $\mathbb{Q}$. Then we can simultaneously obtain the maximal label-irrelevant information $G_E$, and  achieve the complementary subgraph to $G_E$, i.e., $G_I = G_E^-$, which allows further refinement and  environment-invariance interactions. 

\subsection{Environment-invariance soft causal interaction}
\textbf{Motivation.} It is observed that the only  invariance across environments cannot sufficiently contribute to final property as some environments can be taken as the conditions and account for  some molecular properties, which has been illustrated in Fig.~\ref{fig:motiva}. Thus only exploiting the invariance for generalization is limited in information loss.  In this subsection, we propose to refine the environment and enable interactions between extracted environment and invariance. Specifically, we argue that the environment-invariance interactions should include three crucial technical issues, 
\begin{itemize}
	\item Ensuring sufficient mutual  interactions between environment and causal invariant substructures.
	\item The interpretability during learning and aggregation process.
	\item The dimension alignment for easy-to-implement interaction. 
\end{itemize}

\textbf{Solution.} To systematically address above issues, we propose our Soft Causal Interaction (SCI) scheme. In order to capture the interactions between environment representation and invariant causal representation on graph, we  allow partial associated environment aggregating with causal invariance in a learnable manner in our SCA. As shown in Fig.~\ref{fig:OV-SCA}, our SCA consists of two parts, i.e., an Environment-invariance Cross Attention to capture the potential correlations within environment itself and then between environment and invariance. And a Gated Causal Bridge, is designed to dynamically allow the sufficient information injection and  interactions between invariances and associated environment to boost the transfer capacity. 

\begin{algorithm}[h]
	\caption{The training process of CauEMO}
	\label{alg:Algorithm}
	\begin{algorithmic}
		\STATE {\bfseries Input:} graph dataset $ \mathcal{G}$ 
		\STATE {\bfseries Initial:} Functional group set $\mathcal{G}_I = \{ G_i \}$, non-deterministic group set $\mathcal{G}_{N}=\{G_n\}$, the number of epochs $K$.
		\FOR{$i=1$ {\bfseries to} $K$}
		\STATE \bf{Knowledge-enhanced environment generator:}
		\STATE  $\mathcal{G}  \leftarrow \{ G_{new}\} = {\rm Concat}[G_0; G_j] $, where ${G_j} \in \mathcal{G}_{N}$
		\STATE  $p_{uv}  =	{\rm MLP}(h_u, h_v; \bm{W}_{g})$
		\STATE  $\alpha_{u,v} \sim {\rm Bern}(p_{uv})$, ${G_E} \sim \bm{A}_S = \alpha \cdot \bm{A}$
		\STATE \bf{E-GIB environment disentanglement:}
		\STATE  $\bm{Z}_e = \mathbb{P}_\theta({G_E})$, $\bm{Z}_c= \mathbb{P}_\theta({G_E^-})$
		\STATE  $\bm{Z}_e^Q = {\bm{Z}_e}{\bm{W}^Q},\;\bm{Z}_c^K = {\bm{Z}_c}{\bm{W}^K},\;Z_c^V = {\bm{Z}_c}{\bm{W}^V}$
		\STATE  ${ \bm{Z}_e} = {\rm Softmax} (\frac{{ \bm{Z}_e^Q{{( \bm{Z}_c^K)}^T}}}{{\sqrt d }}) \bm{Z}_c^V + \varepsilon $
		\STATE  ${ \bm{Z}_c} = {\rm Softmax} (\frac{{ \bm{Z}_c^K{{( \bm{Z}_e^Q)}^T}}}{{\sqrt d }}) \bm{Z}_c^V + \varepsilon $
		\STATE  Implement invariance and environment interactions with Gated Causal Bridge (Equation.~\ref{eq:EIINter}).	
		\STATE $\widehat{Y} = {\rm MLP}_{{\psi}} (\bm{Z}_{ce})$
		\STATE \bf{Learning optimizing:}
		\STATE ${\rm min} \;\; -\mathbb{E}[log \mathbb{P}_\psi(Y|G)] + \mathbb{E}[log \mathbb{P}_\theta(Y|G_E)]- \beta \mathbb{E}[{\rm KL}({P_\phi }({G|{G_E}})||\mathbb{Q}( {G_E})]$
		\ENDFOR
		\STATE \textbf{Return} $ \psi$, $ \theta$, and $\phi$.
	\end{algorithmic}
\end{algorithm}

In detail, our Environment-invariance Cross Attention is composed of three learnable parameters, $\bm{W}^Q, \bm{W}^K, \bm{W}^V$, we feed the representation of environment $G_E (\bm{Z}_e)$ and causal representation $\bm{Z}_c$ into the attention mechanism. The three hidden representation for soft score calculation can be derived, 
\begin{equation}
	\bm{Z}_e^Q = {\bm{Z}_e}{\bm{W}^Q},\;\bm{Z}_c^K = {\bm{Z}_c}{\bm{W}^K},\;Z_c^V = {\bm{Z}_c}{\bm{W}^V}
\end{equation}
Then we can impose the cross attention to capture the mutual interactions thus obtaining the environment representation $ \bm{Z}_e $ and causal invariance representation $\bm{Z}_c$,
\begin{equation}
	{ \bm{Z}_e} = {\rm Softmax} (\frac{{ \bm{Z}_e^Q{{( \bm{Z}_c^K)}^T}}}{{\sqrt d }}) \bm{Z}_c^V + \varepsilon 
\end{equation}
\begin{equation}
	{ \bm{Z}_c} = {\rm Softmax} (\frac{{ \bm{Z}_c^K{{( \bm{Z}_e^Q)}^T}}}{{\sqrt d }}) \bm{Z}_c^V + \varepsilon 
\end{equation}
where the random noise satisfying $\varepsilon \sim \mathcal{N}(0,\bm{I})$ is added to boost the robustness of representations. Then the learnable coefficient $\{{\rm Softmax} (\frac{{ \bm{Z}_e^Q{{( \bm{Z}_c^K)}^T}}}{{\sqrt d }}), {\rm Softmax} (\frac{{ \bm{Z}_c^K{{( \bm{Z}_e^Q)}^T}}}{{\sqrt d }})\}$ between 0 and 1 are considered as the correlations. 

By obtaining the correlation enhanced representation, we further introduce a Gated Causal Bridge to allow partial relevant environments to be aggregated with invariance substructure representations. It is followed by three steps, 
\begin{itemize}
	\item Dimension alignment between environment and invariant substructure for the generation of our gate where the dimension of $\bm{Z}_e$ is aligned to the same with  $\bm{Z}_c$.
	\item Absorbing the mutual information  to update the respective  $\bm{Z}_e$ and $\bm{Z}_c$ with interpretability.
	\item Implementing the interactions between  $\bm{Z}_e$ and $\bm{Z}_c$ for achieving final interacted representation  $\bm{Z}_{ce}$.
\end{itemize}
We can formulate the above steps in following equations, 
\begin{equation}
	\left\{ \begin{array}{l}
		{\bm{f}_{gate}} ={\rm tanh} ({\bm{W}_{gate}}{\bm{Z}_e})\\
		{\bm{Z}_e} = {\bm{Z}_e} + {\bm{Z}_c} \odot {\bm{f}_{gate}}\\
		{\bm{Z}_c} = {\bm{Z}_c} + {\bm{Z}_e} \odot {\bm{f}_{gate}} \\
		{\bm{Z}_{ce}} = {\bm{Z}_c} \odot {\bm{f}_{gate}} 
	\end{array} \right.
	\label{eq:EIINter}
\end{equation}
where $\odot$ denotes element-wise product, the activation function ${\rm tanh}$ allows both positive and negative signs for the environment output, satisfying the information filtering from environment to invariance. The $\bm{f}_\mathit{gate}$ can be viewed as the squashed environment representation, and the updated $\bm{Z}_c$ can be considered as incorporating the partial relevant environment representation with a soft weighted parameter $\bm{f}_\mathit{gate}$ for final prediction. The detailed analysis regarding interpretability can be found in our case study of  Experiment section.

\subsection{OOD prediction stage}
As the out-of-distribution molecules $\mathbb{G}_{test}=\{G_{t_1}, G_{t_2}, ?\}$ come, we can feed the new graph $G_{t_i}$ into the molecular learning framework CauEMO. Following the irrelevant environment disentanglement, and environment-invariance soft causal interaction, CauEMO can efficiently disentangle the environment part $\bm{Z}^{t_i}_e$ on $G_{t_i} $ and further boost the causal invariance into $\bm{Z}_c$ with a cross attention and environment gate. Then we can take the environment-enhanced causal invariance representation with well soft aggregation $\bm{Z}_{ce}$ for final prediction, i.e.,
\begin{equation}
	\widehat{Y} = {\rm MLP}_{\psi} (\bm{Z}_{ce})
\end{equation}
where ${\psi}$ are parameters for final classifier or regressor. 

\begin{table*}
	\caption{OOD generalization performance (ROC-AUC). The best results are in \textbf{bold}  and the second best is \underline{underlined}.}
	\centering
	\resizebox{0.95\linewidth}{!}{
	\begin{tabular}{cccccccccc}
		\hline 
		& {EC50-Assay}   & {EC50-Sca}   &{EC50-Size}   & {Ki-Assay}    & {Ki-Sca}  & {Ki-Size} \\  \hline
		GCN        & 61.20$\pm$ 1.60    & 65.3 $\pm$ 1.91     & 52.1 $\pm$ 2.0     & 83.7 $\pm$ 4.70  & 33.2 $\pm$ 1.81   & 31.6 $\pm$ 1.72  \\ 
		\tabincell{c}{Graph-\\SAGE}        & 74.8 $\pm$ 3.40    & 64.1 $\pm$ 2.84     & 52.5 $\pm$ 1.63    & 84.6 $\pm$ 5.32   & 34.8 $\pm$ 2.00   & 31.5 $\pm$ 2.50   \\
		GIN         & 75.8 $\pm$ 1.31    & 66.4 $\pm$ 2.00     & 56.2 $\pm$ 1.60    & 89.4 $\pm$ 5.62    & 39.9 $\pm$ 1.31   & 39.0 $\pm$ 1.60     \\ \hline
		\tabincell{c}{IB-\\subgraph}    & 75.31$\pm$2.06    & 62.91$\pm$1.67     & 60.57$\pm$2.03    & 72.41$\pm$1.23    & 70.67$\pm$2.30   & 73.65$\pm$2.34   \\ 
		GSAT       & 76.07$\pm$1.95    & {63.58$\pm$1.36}     & 61.12$\pm$0.66     & {72.26$\pm$1.76}  & 71.16$\pm$0.80   & {75.78$\pm$2.60}   \\
		DIR    & 74.51$\pm$2.12    & 63.23$\pm$1.44     & 61.82$\pm$1.04     & 71.92$\pm$1.23  & 69.56$\pm$0.43    & 74.98$\pm$1.96     \\ 
		CIGA    & {75.03$\pm$2.47}     & {65.41$\pm$1.16}     & {64.10$\pm$1.08}     & {73.95$\pm$2.50}  	   & 71.87$\pm$3.32   & {74.46$\pm$2.32} 	  \\ 
		GALA    & {77.56$\pm$2.88}     & \underline{66.28$\pm$0.45}     & \underline{64.25$\pm$1.21}     & \underline{77.92$\pm$2.48}  	   & 73.17$\pm$0.88   & \underline{77.40$\pm$2.04} 	  \\ 
		IGM    & \underline{77.60$\pm$2.10}     & {65.74$\pm$1.04}     & {63.45$\pm$1.19}     & {75.63$\pm$2.40}  	   & \underline{73.83$\pm$1.60}   & {76.95$\pm$2.19} 	  \\  \hline
		\tabincell{c}{CauEMO\\(Ours)}    & \textbf{78.46$\pm$1.42}     & \textbf{66.91$\pm$0.62}     & \textbf{65.77$\pm$1.03}     & \textbf{78.12$\pm$1.81}  	   & \textbf{74.53$\pm$1.10}   & \textbf{78.21$\pm$1.92}    \\ \hline
		\label{tab:result_2}
	\end{tabular}}
\end{table*}

\section{Experiment}
We evaluate CauEMO using both synthetic and real-world datasets by explicitly involving distribution shifts. Both practices of causal invariance  and  environment-based methods are taken for comparison. Specifically, we would like to answer  the following two questions via empirical experiments: 
\begin{itemize}
	\item  On scenarios where environmental information dominates the graph, can our CauEMO outperform existing methods? 
	\item  When the associations between invariant rationales and labels are implicit, can CauEMO capture true causal associations?
\end{itemize}

\subsection{Dataset}
The  datasets for evaluation are three-fold. We choose totally 7 datasets, including four real-world molecular datasets, two categories of drugOOD  datasets and one synthetic dataset of Motif to verify the effectiveness of CauEMO.

\textbf{Real-world datasets.} 
There are \textbf{four} real-world datasets on molecular property prediction. 
\begin{itemize}
	\item  \textbf{MUTAG}~\cite{debnath1991structure} is a binary dataset of molecular property, where nodes indicate atoms and edges denote chemical bonds. Each graph is associated with a binary label based on its mutagenic effect.
	\item \textbf{Open Graph Benchmark (OGB)}~\cite{hu2020open} is a series of real, large-scale and diverse datasets which are utilized for machine learning on graphs. It covers almost all real-world tasks, including node-level, link-level and graph-level property prediction. We choose \textbf{MOLHIV}, \textbf{BBBP} and \textbf{SIDER} to verify our method.
\end{itemize}

\textbf{DrugOOD datasets.} 
To evaluate the OOD performance of CauEMO, we adopt 6 sub-datasets from two categories  of \textbf{DrugOOD benchmark}~\cite{ji2022drugood}. It focuses on the challenging real-world task of AI-aided drug affinity prediction. The distribution shift happens on different Assays, Scaffolds and molecule Sizes. In particular, \textbf{DrugOOD-lbap-core-ec50-assay, DrugOOD-lbap-core-ec50-scaffold,  DrugOOD-lbap-core-ec50-size, DrugOOD-lbap-core-ki-assay, DrugOOD-lbap-core-ki-scaffold, and DrugOOD-lbap-core-ki-size} are selected.

\textbf{Synthetic datasets.} We select a synthetic dataset to assiduously verify the validity and interpretability of CauEMO. \textbf{Spurious-Motif} is a synthetic dataset proposed by \cite{wu2022discovering} with three graph classes. Each graph is composed of one base $S$ and one motif $C$. The motif $C$ directly determines the label of the graph. We can create Spurious-Motif datasets with different spurious correlations, which represents the degree ($b$) between the base $S$ and the label. In our implementation, we let $b = 0.5, 0.7, 0.9$ for dataset generation. 
\begin{figure}[ht]	
	\centering
	\includegraphics[width=5.2in]{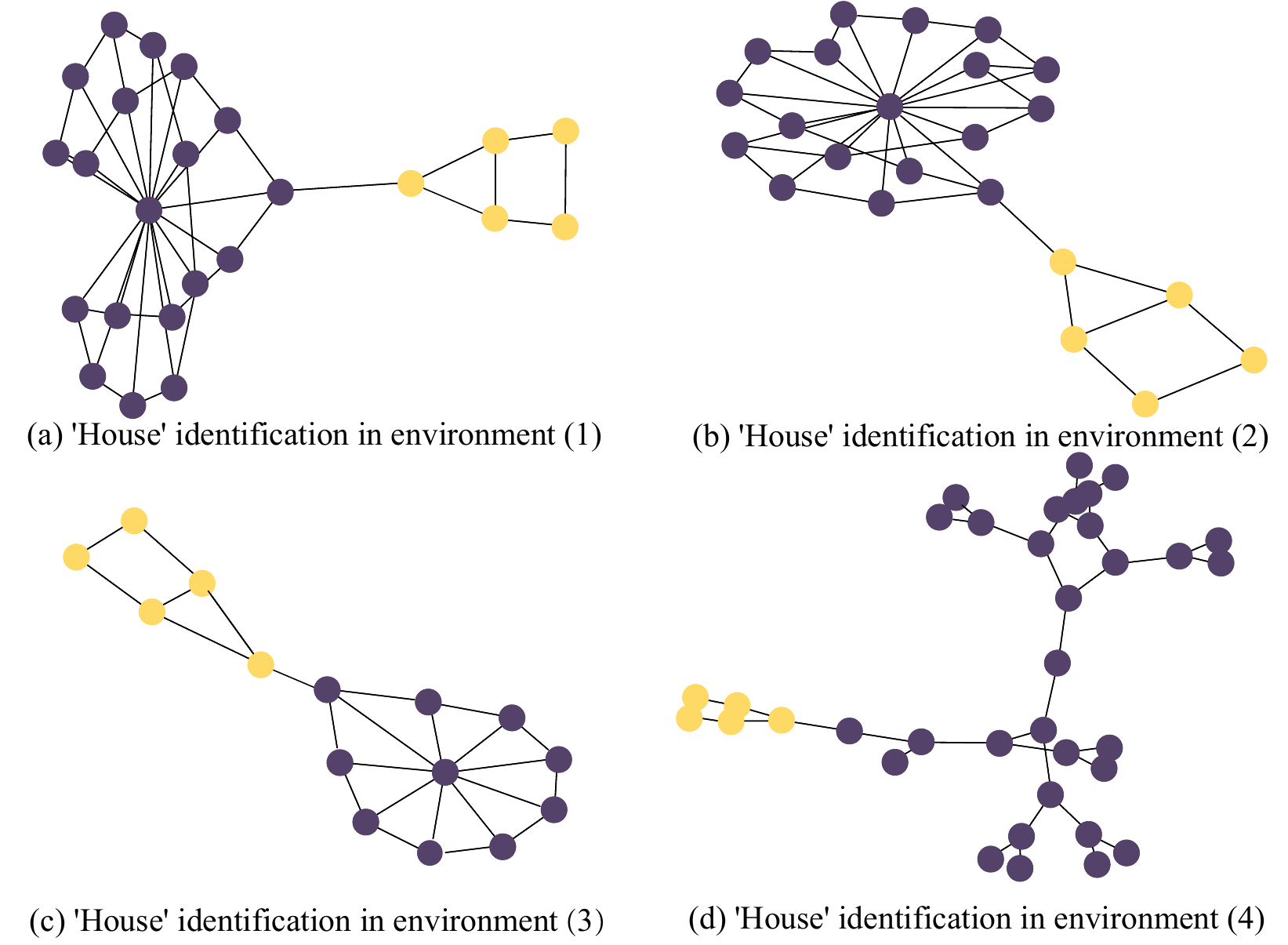}
	\caption{The ability of CauEMO to identify 'house' in Spurious-Motif dataset.}
	\label{fig:motif}
\end{figure}

\begin{figure}[ht]
	\centering
	\subfigure[Performance comparison between CauEMO and CauEMO-Random.]{               
		\includegraphics[scale=0.35]{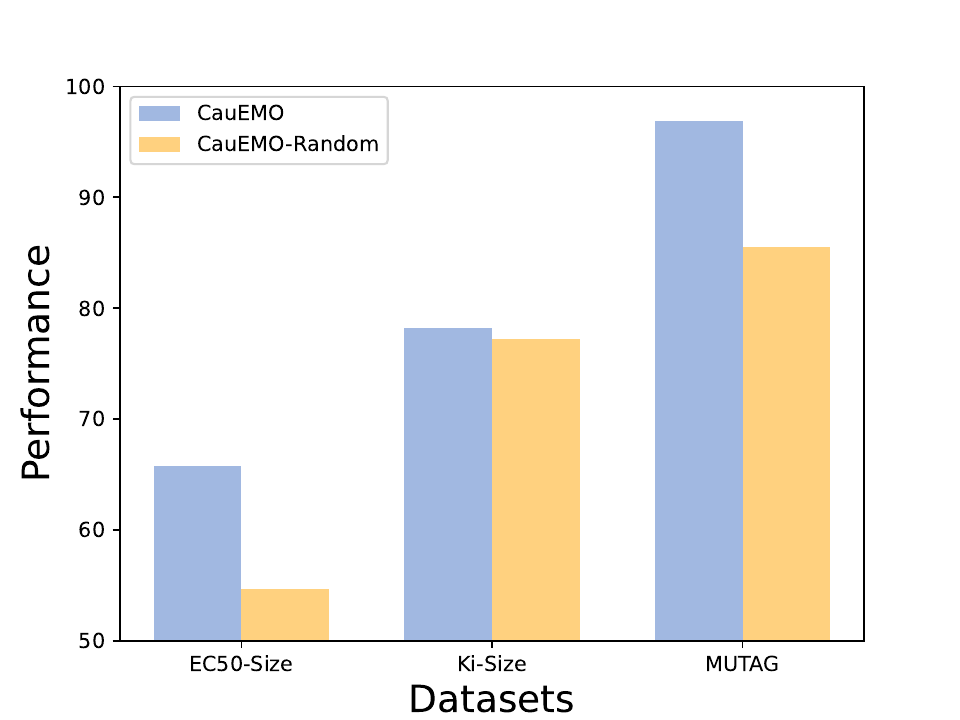}
	}
	\hspace{0in}
	\subfigure[Performance comparison between CauEMO and CauEMO-Subgraph.]{
		\includegraphics[scale=0.35]{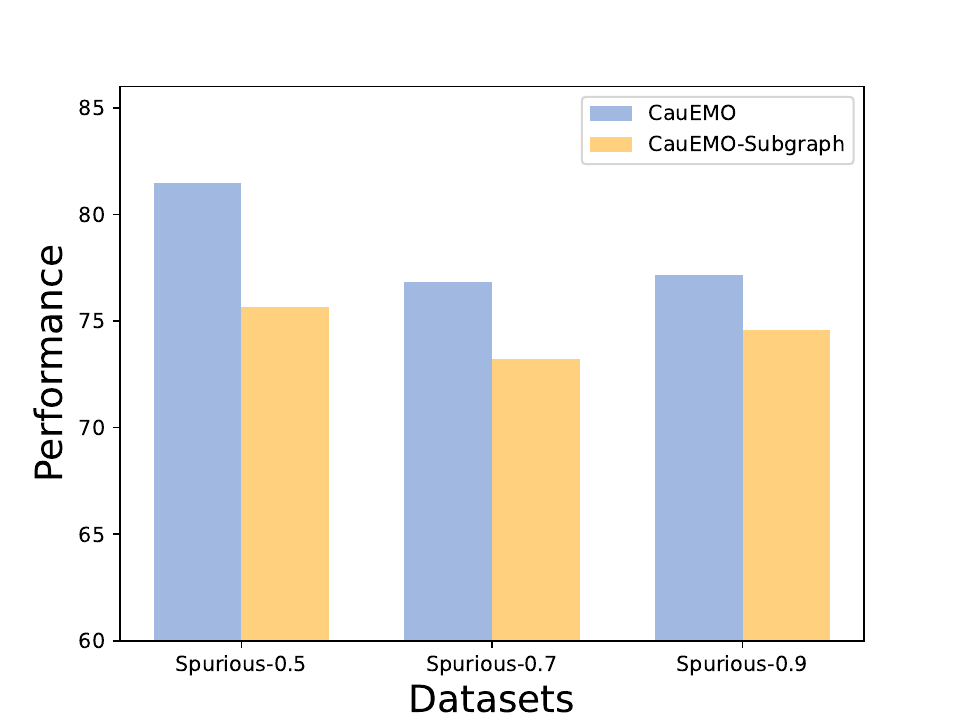}
	}
	\caption{Ablation studies on CauEMO-Random and CauEMO-Subgraph}
	\label{fig:Ablation_RS}
\end{figure}

\begin{figure*}
	\centering
	\includegraphics[width=5.4in]{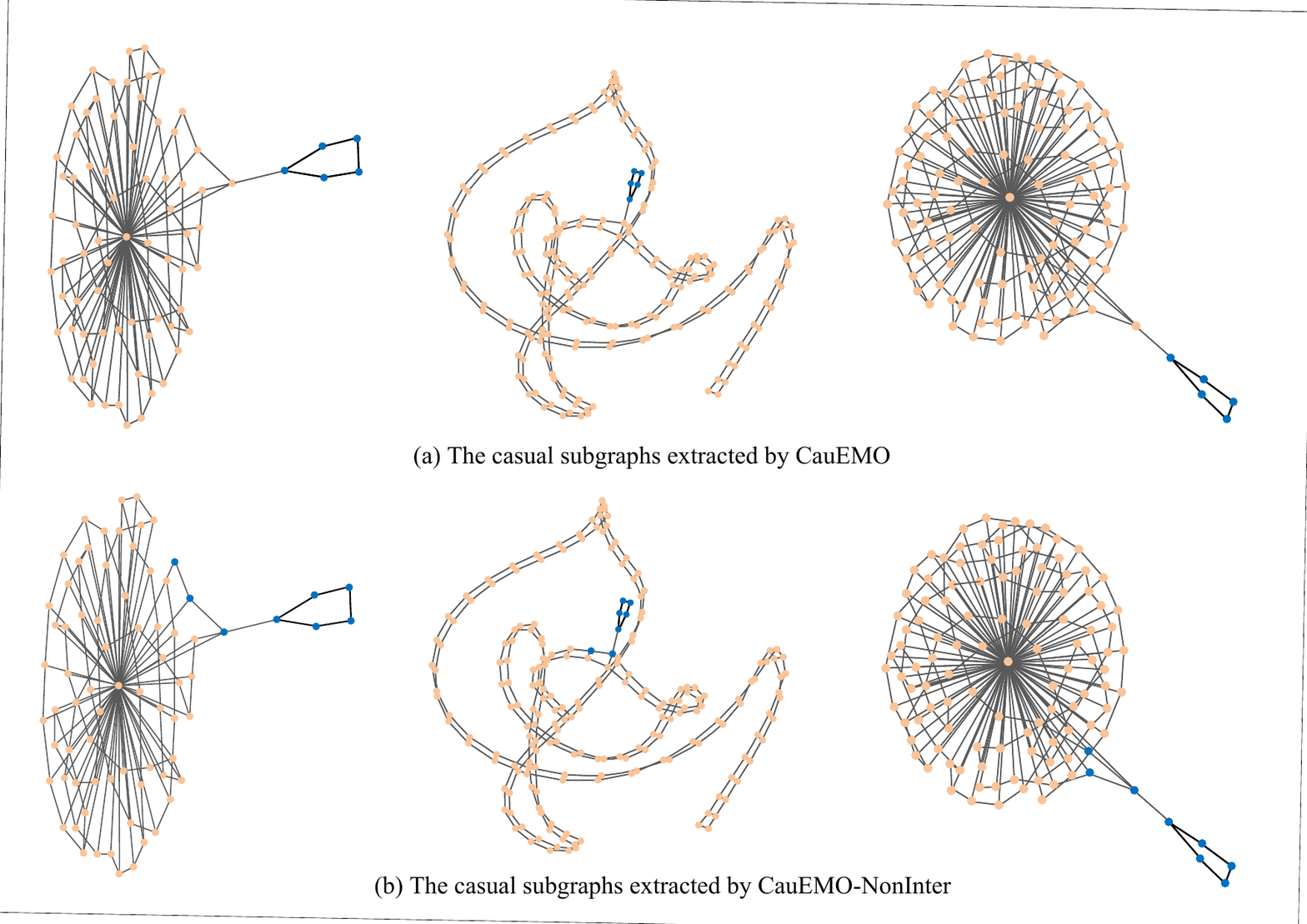}
	\caption{The comparison between CauEMO and its ablative variant without interaction mechanism, CauEMO-NonInter.}
	\label{fig:visual}
\end{figure*}

\begin{figure*}[ht]
	\centering
	\subfigure[Performance analysis of the hyperparameter $\beta$.]{               
		\includegraphics[scale=0.12]{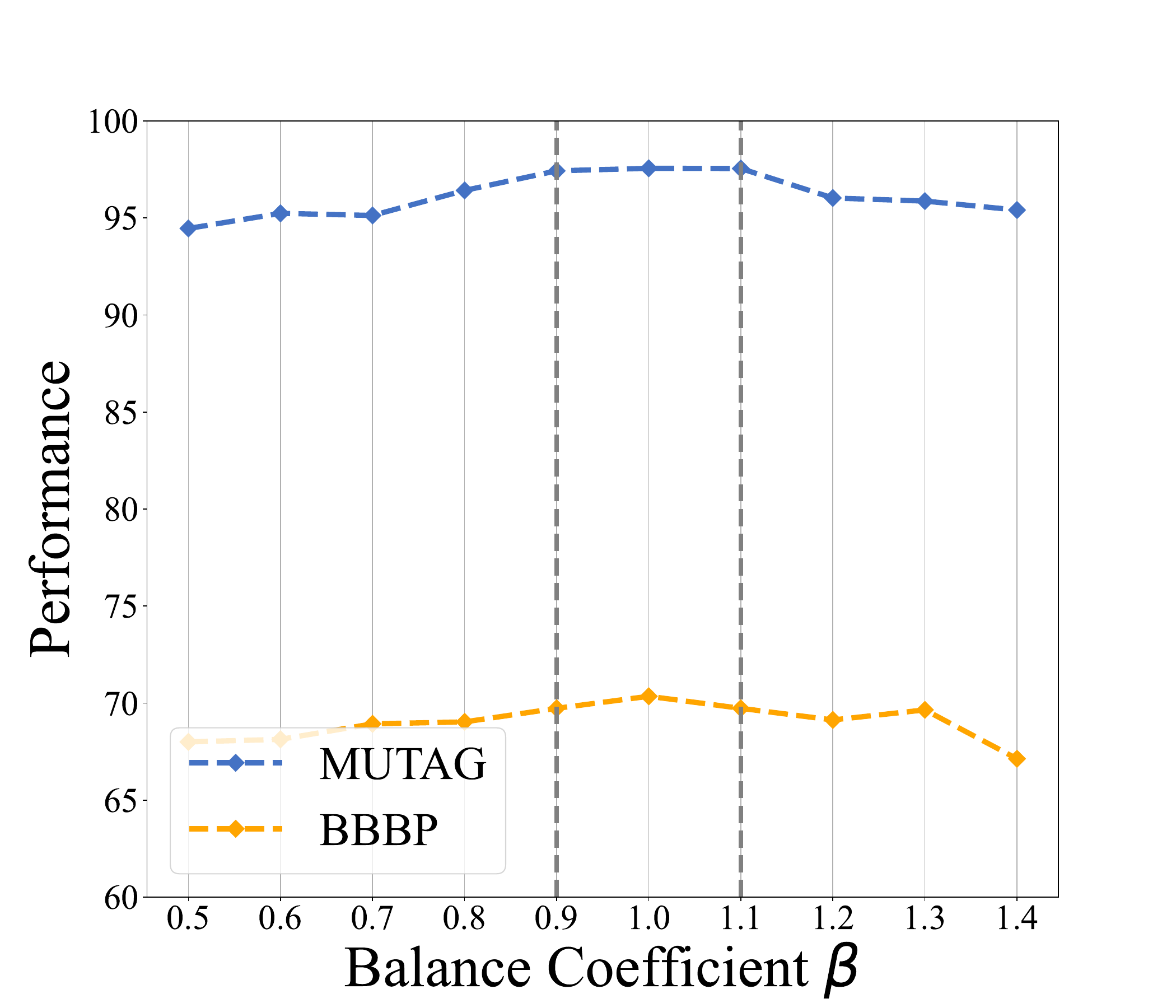}
	}
	\hspace{0in}
	\subfigure[Performance comparison of the dimension of $\bm{Z}_e$.]{
		\includegraphics[scale=0.30]{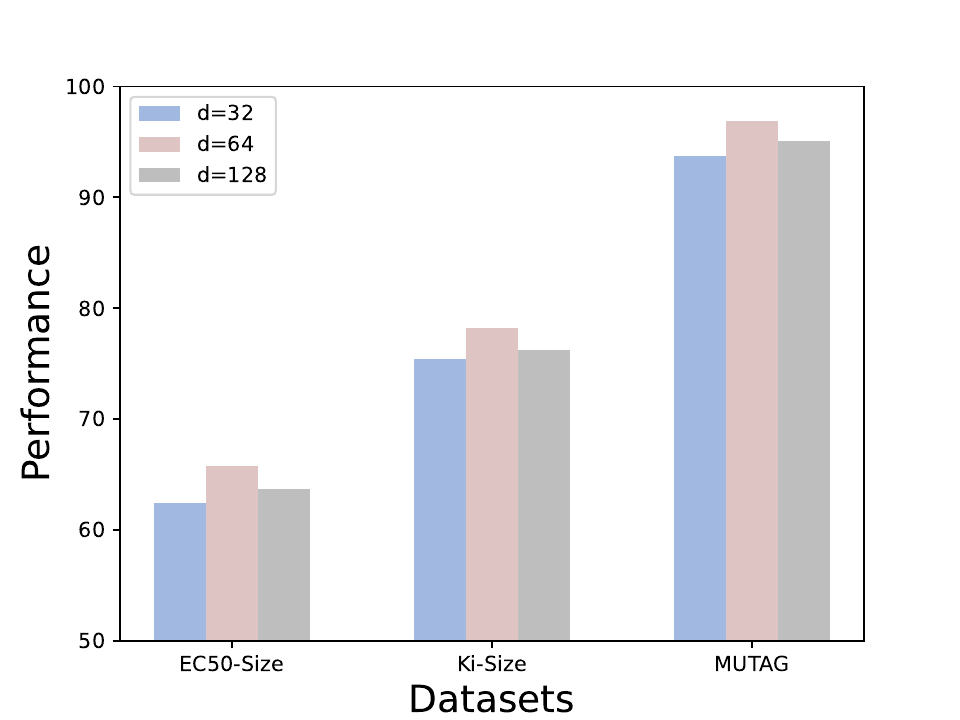}
	}
	\hspace{0in}
	\subfigure[Performance comparison of the dimension of $\bm{Z}_c$.]{
		\includegraphics[scale=0.30]{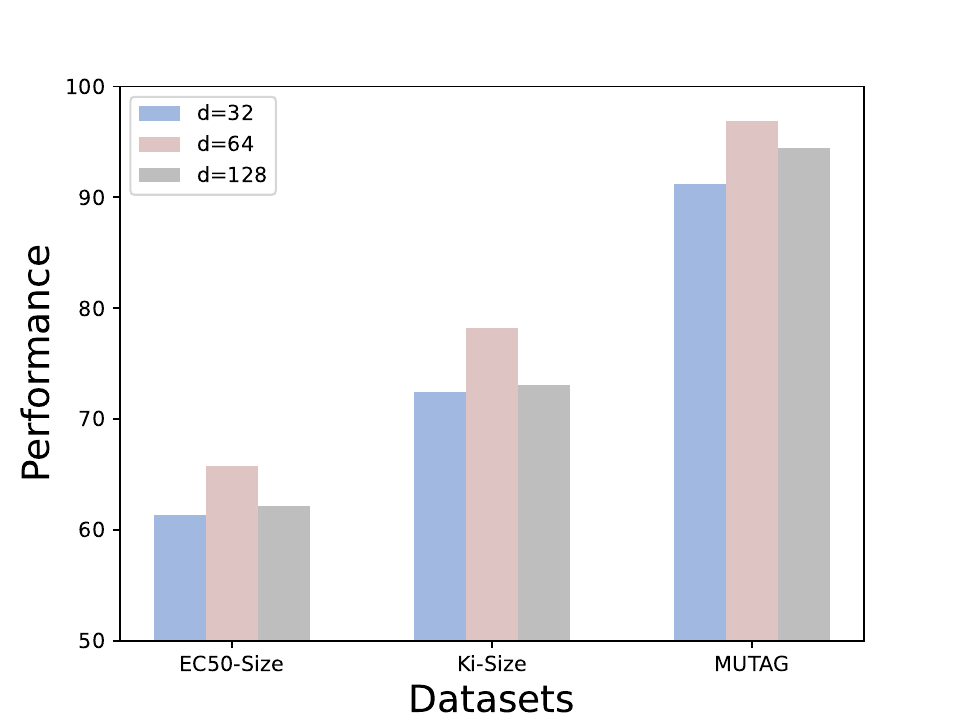}
	}
	\hspace{0in}
	\subfigure[Performance comparison of the dimension of $\bm{Z}_{ce}$.]{
		\includegraphics[scale=0.30]{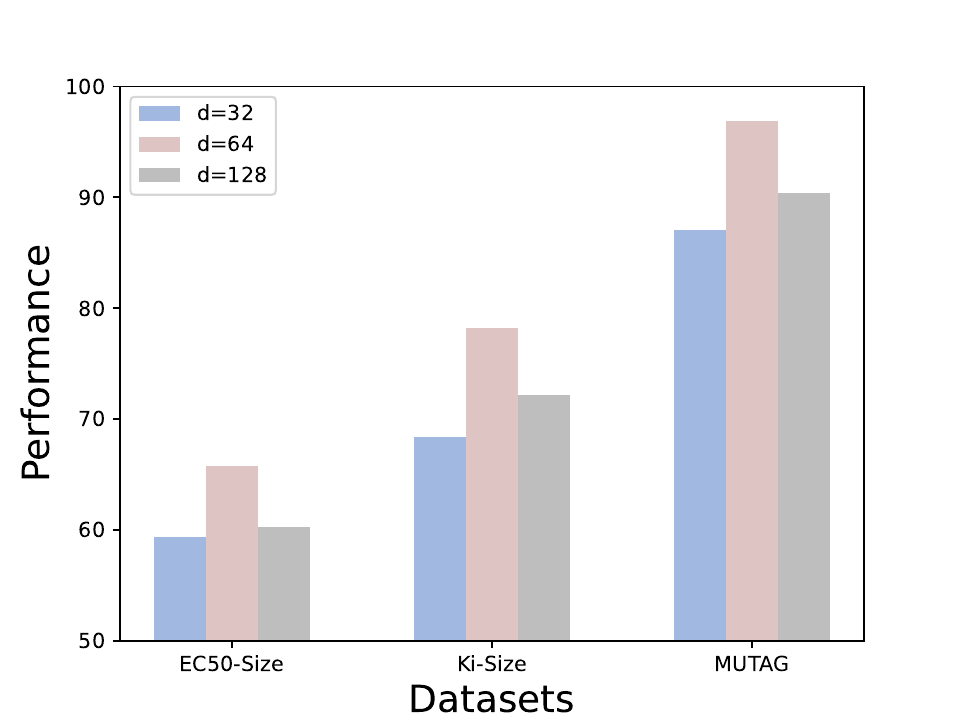}
	}
	\caption{Performance variations during hyperparameter adjustment}
	\label{fig:hyper}
\end{figure*}

\subsection{Baselines}
We choose three categories of baselines, including conventional GNN backbones, subgraph-based invariant learning methods and environment-based graph learning models. 

\textbf{Conventional backbone baselines.} Three popular backbones in most practices, are taken as our baselines for evaluation.
\begin{itemize}
	\item \textbf{GCN}~\cite{kipf2016semi} is a vanilla Graph Convolution Neural Network via capturing the spatial adjacency for aggregating neighborhood.
	
	\item \textbf{Graph-SAGE}~\cite{hamilton2017inductive} takes a random neighbor sampling strategy to simplify the computation of information fusion and it allows inductive learning on new nodes.
	
	\item \textbf{GIN}~\cite{xu2018powerful} is an isomorphism graph network to ensure the consistent structure to be with similar representations.
\end{itemize}

\textbf{Subgraph-based invariant learning methods.} We adopt four typical subgraph-based learning baselines for evaluation. 
\begin{itemize}
	\item \textbf{SUN}~\cite{frasca2022understanding} studies the characteristics of node-based subgraph learning and aligns the permutation group of nodes and subgraphs, modeling the symmetry with a smaller single permutation group. 
	\item \textbf{IB-subgraph}~\cite{yu2020graph} first implements the information bottleneck  with graph learning, which is not only a subgraph learning based on partition (edge drop) but also an important exploration of interpretability. 
	\item \textbf{GSAT}~\cite{miao2022interpretable} follows this practice and designs a subgraph extraction strategy with edge deletions based on stochastic attention mechanism.
	\item \textbf{DIR}~\cite{wu2022discovering} splits the input graph into causal and non-causal subgraphs, and utilizes invariant features to construct interpretable model.
\end{itemize} 

\textbf{Environment-based graph learning models.} We exploit three baselines with  explicitly considering the modeling of graph environments. All of them focus on the graph-level out-of-distribution generalization from the perspective of environment modeling.
\begin{itemize}
	\item \textbf{CIGA} \cite{chen2022learning} is an environment-base learning architecture, which utilizes contrastive learning within the same class labels, and assume samples with the same label share invariant substructures. 
	\item \textbf{GALA}~\cite{chen2023does} is a  symmetric graph convolutional autoencoder for unsupervised graph representation learning. 
	\item \textbf{IGM}~\cite{jia2024graph} proposes to exploit the environment to augment the learning of invariance. 
\end{itemize}

\begin{table*}[ht]
	\caption{Performance comparison. We report ROC-AUC for MOLHIV, BBBP, and SIDER. Besides, we also report Accuracy on MUTAG and Spurious-Motif for evaluation. The best results are in \textbf{bold}  and the second best is \underline{underlined}.}
	\centering
	\resizebox{0.95\linewidth}{!}{
	\begin{tabular}{ccccccccc}
		\hline 
		& \multirow{2}{*}{MOLHIV}   & \multirow{2}{*}{BBBP}   &\multirow{2}{*}{SIDER}   & \multirow{2}{*}{MUTAG}    & \multicolumn{3}{c}{Spurious-Motif}  \\ 
		& \multirow{2}{*}{}       & \multirow{2}{*}{}   & \multirow{2}{*}{}    &\multirow{2}{*}{}      & 0.5  &  0.7    & 0.9           \\  \hline
		GCN        & 75.52$\pm$1.61    & 65.34$\pm$1.94     & 52.12$\pm$2.03     & 83.75$\pm$4.74  & 33.22$\pm$1.82   & 31.68$\pm$1.73     & 29.61$\pm$6.23      \\ 
		Graph-SAGE        & 74.82$\pm$3.42    & 64.16$\pm$2.83     & 52.52$\pm$1.69   & 84.60$\pm$5.34   & 34.88$\pm$2.08   & 31.53$\pm$2.50     & 30.42$\pm$3.47       \\
		GIN         & 75.86$\pm$1.30    & 66.46$\pm$2.00    & 56.24$\pm$1.64    & 89.42$\pm$5.63    & 39.96$\pm$1.35   & 39.04$\pm$1.60    & 38.62$\pm$2.33             \\  \hline
		IB-subgraph    & 76.43$\pm$2.65    & 68.12$\pm$1.12    & 57.71$\pm$2.14    & 94.33$\pm$6.44    & 54.36$\pm$7.09  & 48.51$\pm$5.76    & {46.19$\pm$5.63}             \\ 
		GSAT       & 76.47$\pm$1.53    & {69.13$\pm$2.02}     & \underline{59.19$\pm$1.03}   & {96.37$\pm$2.15}  & 52.74$\pm$4.08   & {49.12$\pm$3.29}     & 44.22$\pm$5.57      \\
		DIR    & 76.34$\pm$1.01    & 69.73$\pm$1.54     & 58.81$\pm$1.84     & 96.01$\pm$2.24  & 58.73$\pm$2.15    & 43.36$\pm$1.64    & 39.87$\pm$0.56               \\ 
		CIGA    & {76.94$\pm$1.32}     & {69.65$\pm$1.32}     & {58.95$\pm$1.22}     & {95.77$\pm$1.23}  	   & 77.33$\pm$9.13   & {69.29$\pm$3.06} 	  & {63.41$\pm$7.38}        \\ 
		GALA    & {77.04$\pm$1.60}     & {70.21$\pm$1.31}     & {59.04$\pm$1.30}     & \underline{96.76$\pm$1.70}  	   & 73.45$\pm$5.43   & {68.56$\pm$3.32} 	  & {69.82$\pm$2.34}        \\ 
		IGM    & \underline{77.20$\pm$1.39}     & \underline{71.03$\pm$0.79}     & {58.23$\pm$1.43}     & {96.04$\pm$2.01}  	   & \textbf{82.36$\pm$7.39}   & \textbf{78.09$\pm$5.63}	  & \underline{76.11$\pm$8.86}        \\  \hline
		\tabincell{c}{CauEMO\\(Ours)}     & \textbf{77.91$\pm$1.54}     & \textbf{72.31$\pm$1.02}     & \textbf{59.90$\pm$1.28}     & \textbf{96.92$\pm$1.36}  	   & \underline{81.45$\pm$6.05}   & \underline{76.82$\pm$2.81} 	  & \textbf{77.17$\pm$5.95}        \\ \hline
		\label{tab:result_1}
	\end{tabular}}
\end{table*}

\begin{table}[ht]
	\caption{The ablative performance comparison between CauEMO and CauEMO-NonGCB.}
	\centering
	\begin{tabular}{ccccccccc}
	\hline 
	& \multirow{2}{*}{MOLHIV}   & \multirow{2}{*}{BBBP}   &\multirow{2}{*}{SIDER}   & \multirow{2}{*}{MUTAG}    & \multicolumn{3}{c}{Spurious-Motif}  \\ 
	& \multirow{2}{*}{}       & \multirow{2}{*}{}   & \multirow{2}{*}{}    &\multirow{2}{*}{}      & 0.5  &  0.7    & 0.9           \\  \hline
	CauEMO                     & 77.91    & 72.31    & 59.90    & 96.92  & 81.45  & 76.82    & 77.17  \\ \hline
	\tabincell{c}{CauEMO\\-NonGCB}& 75.52  & 70.14   & 54.02    & 93.37   & 76.58 &  72.41 & 73.62  \\ \hline
	\label{tab:Ablation_1}
	\end{tabular}
\end{table}

\subsection{Evaluation metrics and implementation details}
We employ the same metrics as the previous approach to evaluate specific dataset. For the MUTAG and Spurious-Motif datasets, we exploit \textbf{accuracy} as the evaluation metric.

For the DrugOOD and OGB datasets, we evaluate the performances using the \textbf{ROC-AUC} metric where the value of this metric is the higher, the better. We report the mean results and standard deviations across ten runs. We exploit GIN as the backbone of CauEMO and all experiments are conducted on an NVIDIA A100-PCIE-40GB.

\subsection{Performance comparison}

\textbf{OOD generalization performance under distribution shifts.} In Table \ref{tab:result_2}, we report the ROC-AUC on six distribution-shift datasets. We can clearly observe that our CauEMO consistently achieves the best performance across five datasets. This demonstrates that our environment-centered design can achieve superior performance under distribution shift scenarios. Moreover, we also have the following two observations. 
1) Compared with conventional backbone GNNs, those graph learning models specially designed for OOD scenarios have better performance. This explicitly confirms the validity and rationality of existing  invariant learning methods and environment-based models. 2) Compared with the methods based on causal invariance theory, the environment-oriented models perform better. GALA and IGM obtain sub-optimal results across all datasets, suggesting these solutions can potentially improve OOD generalization capacity with environment-oriented strategy. This serves as the practical foundation of our work on explicit environment modeling and disentanglement.

\textbf{Prediction performance of real-world tasks and interpretability.} In Table \ref{tab:result_1}, we show the prediction performance of CauEMO on four real-world datasets and three synthetic datasets. The results suggest that CauEMO achieves competitive performance in real-world molecular classification tasks, reaching five  best performances across  seven datasets.
Noted that IGM, which utilizes the cooperative mix-up strategy combining both environment and invariance parts, slightly outperforms our CauEMO on two datasets and it can potentially verify that the intuition of environment-invariance cooperation makes sense. It is worth noting that CauEMO has gained more capacity on generalization over other five datasets, which may be attributed to the sufficient mutual interactions in environment-invariance SCA, and label-irrelevant information squash of E-GIB. We choose the synthetic dataset to explore whether CauEMO could identify specific causal substructures. As shown in Fig.\ref{fig:motif}, we present the ability of CauEMO to discover the structure of 'house' around various environments in Spurious-Motif dataset. Despite the diversity of surrounding four environments in Motif, our CauEMO  can always accurately identify the invariant property substructures and we believe the potential reason behind the superior performances derived from  perceiving broader various environments that is complementary to invariances.

\subsection{Ablation study}
We conduct ablation studies from following four aspects: 
\begin{itemize}
	\item Whether chemistry-guided environment generator can benefit uncover environment diversity? 
	\item Whether the environment-centered graph information bottleneck design is superior to an invariant subgraph-centered approach?
	\item Whether the interaction mechanism between invariant subgraphs and environment subgraphs through cross-attention in soft causal-invariance interaction is beneficial to performance improvement?
	\item Whether the design of a Gated Causal Bridge is helpful for isolating environment variables while preserving the causal invariant subgraph? 
\end{itemize}

Firstly, we  design a variant of CauEMO, CauEMO-Random, which utilizes random noise to achieve environment growth. We conduct experiments on EC50-Size, Ki-Size and MUTAG datasets. Fig.~\ref{fig:Ablation_RS}(a) shows the comparison between CauEMO and CauEMO-Random. We observe that CauEMO-Random achieves a worse performance than CauEMO on most datasets.  This indicates that environment variables in molecules show strong domain-specific characteristics, and we can conclude that incorporating environment information into molecules in a random and unstructured way often causes shifts in their inherent properties. Therefore, our designed chemistry-guided environment generator can effectively enhance the discovery of environmental diversity by introducing structured and contextually relevant variations, allowing for a more comprehensive exploration of diverse environmental conditions.

Secondly, we propose a variant  centered on causal subgraph learning, CauEMO-Subgraph. This variant does not focus on modeling the environment factors but instead remains aimed at extracting causal subgraphs. On the Spurious-Motif datasets, we compare the performance of CauEMO and CauEMO-Subgraph. As shown in Fig.~\ref{fig:Ablation_RS}(b), we observe a significant performance degradation in CauEMO-Subgraph, highlighting the effectiveness of our environment-centered approach. We argue that this performance degradation stems from the fact that positive causal learning often fails to isolate invariant subgraphs in complex environments. Modeling the environment factors, which effectively incorporate both positive and negative learning, provides a more robust solution for OOD generalization in graph learning.

Thirdly, we aim to verify whether the interaction between causal variables and environment variables in graphs can enhance model robustness. To this end, we propose a variant without an interaction mechanism, CauEMO-NonInter, where the learned environment representation does not contribute to enhancing the positive invariant learning process. As shown in Fig.~\ref{fig:visual}, we can observe that the invariant subgraphs learned by CauEMO-NonInter are usually with several environment nodes those erroneously identified (Second line of Fig.~\ref{fig:visual}). In contrast, our CauEMO can achieve clearer extractions of invariant subgraphs. This can potentially indicate that the interaction of causal variables and environment variables makes environment-invariance easier to separate.

Finally, we aim to validate the effectiveness of the Gated Causal Bridge. Thus, we can obtain a variant of CauEMO, CauEMO-NonGCB, by removing the Gated Causal Bridge component. As shown in Table \ref{tab:Ablation_1}, the results on all datasets reveal that the ablative variant is inferior to our CauEMO (integrated one).  These empirical results further confirm that the design of Gated Causal Bridge is beneficial to enhance the effectiveness and robustness of the model.

\subsection{Hyperparameter analysis}
The important hyperparameters in our study are two-fold. First, in E-GIB, the balance parameter $\beta$ in Equation.(\ref{eq:EGIB}) plays the role of balancing the trade-off of compositional invariance and environment. We set $\beta$ in the interval $[0.5,1.4]$ and visualize the performances  on two selected real-world datasets MUTAG and BBBP for generalization task. Second, in soft causal interaction (SCI), the latent dimension of representation, i.e., molecular environment representation ($\bm{Z}_e$), molecular invariance representation ($\bm{Z}_c$) and gated environment-invariance representation ($\bm{Z}_{ce}$), where the dimensions are ranging in the scale $\{\mathbb{R}^{16 \times 1}, \mathbb{R}^{32 \times 1}, \mathbb{R}^{64\times 1}, \mathbb{R}^{128\times 1}\}$. The larger dimension may indicate more learning capacity while simultaneously means more complexity and computational workloads. We also perform such dimension-related sensitivity tests on datasets of MUTAG and BBBP
on  generalization task. We visualize our hyperparameter analysis process in Fig.~\ref{fig:hyper}.
Regarding $\beta$, it experiences a climbing stage and a downward stage where it achieves best performances during the interval $[0.9,1.1]$ on both two datasets, then  $\beta$ is set to $1$ in our implementation. For the dimension of representation,  it also experiences a climb and then drop down with dimension increasing, we then set 64 across all datasets as an intermediate trade-off for final experiments. 

\section{Conclusion}
In this work, we propose a novel graph learning framework, CauEMO, to address the OOD challenges in molecule science, from the perspective of environment growth and environment-invariance interactions on graphs. We systematically address such OOD prediction over molecular graph property on three  aspects. First, to extend  the scale of environment and ensure the augmented quality of molecules, a chemistry bond principle-based domain knowledge enhanced molecular environment generator is proposed. Second, for maximumly squashing the irrelevant information from  the whole graph, we devise an Environment-GIB based irrelevant environment disentanglement via deriving a modified environment-based graph information bottleneck, which not only decouples the causal invariant substructures, but also provides the interpretability and theoretical guarantee for our solution. Third, in order to allow sufficient information interactions between extracted environment and causal invarainces,   we further devise an environment-invariance soft causal interaction, which consists of a cross-attention mechanism for weighting the importance of environments and a gated causal bridge to enable dynamical interactions of two branches. We conduct extensive experiments on 7 datasets against  9 baselines including conventional graph learning backbones, subgraph-based backbones as well as environment-based learning backbones. The results on performance comparison and ablation studies demonstrate the overall superiority of our CauEMO and the effectiveness of each module in CauEMO. We believe our CauEMO can be a real interdisciplinary solution that intersects data mining, bio-informatics and chemistry with informative scientific insights. 

In the future, our research plans can be three fold. First, we will further investigate how to discover more causal invariant subgraphs, i.e., more than one substructures in the whole graph, regarding properties from diverse chemistry and bioinformatic  insights. Second, we are going to study how to design the fusion mechanism among multiple substructures as well as environments on graphs, and enable a multi-substructure to multi-property research scheme. Third, regarding  OOD learning, how can we improve the strategy of  invaraince extraction and model adaptation, to allow our model to adaptively adapt the newly arrived data or very few new data to realize the evolution of learning AI models. 	
\section{Acknowledgment}

This paper is partially supported by the National Natural Science Foundation of China (No.12227901),  and Natural Science Foundation of Jiangsu Province (BK.20240460).
%
%
%
%
%
%

\bibliography{References}	

\end{document}